# Towards Discovering Erratic Behavior in Robotic Process Automation with Statistical Process Control

Petr Průcha[0000-0003-2197-7825]

Technical University of Liberec, Liberec, Czechia
petr.prucha@tul.cz

**Abstract.** Companies that use robotic process automation very often deal with problems maintaining the bots in their RPA portfolio. Current key performance indicators do not track the behavior of RPA bots or processes. For better maintainability of RPA bots, it is crucial to easily identify problematic behavior in RPA bots. Therefore, we propose a strategy that tracks and measures the behavior of processes to increase the maintainability of RPA bots. We selected indicators of statistical dispersion for measuring variability to analyze the behavior of RPA bots. We analyzed how well statistical dispersion can describe the behavior of RPA bots on 12 processes. The results provide evidence that, by using statistical dispersion for behavioral analysis, the unwanted behavior of RPA bots can be described. Our results showed that statistical dispersion can describe the success rate with a correlation of -0.91 and outliers in the data with a correlation of 0.42. Also, the results demonstrate that the outliers do not influence the success rate of RPA bots. This research implies that we can describe the behavior of RPA bots with variable analysis. Furthermore, with high probability, it can also be used for analyzing other processes, as a tool for gaining insights into performance and as a benchmark tool for comparing or selecting a process to rework.

## 1 Introduction

The interest in improving processes is as old as humanity itself. Of course, the massive interest in process improvement came with the advent of industrialization and the industrial revolutions. Since then, disciplines have emerged that deal with process improvement. There is also a growing number of scientific publications dealing with processes in the WOS and Scopus databases (Web of Science; Scopus). Naturally, with the development of process sciences, methods of process control and process self-maintenance have also been developing.

Robotic process automation (RPA) is one approach to automating processes (Czarnecki and Fettke, 2021). RPA technology is relatively new, but many companies have already started to use this technology. With new approaches come new problems and RPA is no exception to this rule, bringing new challenges (Kedziora and Penttinen, 2021; Průcha, 2021; Syed et al., 2020; Willcocks et al., 2015a). One of the challenges of RPA according to Syed et al., (2020). is the measurement of RPA performance.

Very often, when measuring the performance of various key performance indicators (KPIs) such as return on investment (ROI), full-time equivalent (FTE) or Run-time of RPA, a number of automations are used (Hofmann et al., 2020). The baseline characteristics do not reveal much about the behavior of the RPA bot or process. Therefore, industry and research studies suggest measuring other KPIs that better describe the behavior of RPA bots (Hofmann et al., 2020; Blueprint, 2021; Teodorczuk, 2021). One of these KPIs is the stability of the RPA bot, not only in terms of RPA bot exceptions, but also in terms of temporal consistency in processing the same cases, i.e., temporal variability in time (Casey, 2019). Looking at the temporal variability of the RPA bot while processing cases is as important as it is in other processes, since a deviation from the norm can indicate an error behavior or an anomaly. There is variability in processes automated by RPA, even if the software involved has precise instructions (Anagnoste, 2017; Sullivan et al., 2021). Deviations from the mean can be measured by statistical dispersion, hence the variability of the RPA bot or process can be measured.

Measuring the variability of an automated process can very well indicate a properly functioning RPA bot, but also a very problematic one. The possibility of selecting and detecting a problematic RPA bot is an activity that simplifies the maintenance of the robot portfolio. As the number of RPA bot deployments increase beyond a hundred automated processes using RPA, detecting problematic behavior starts to exceed the discernment capabilities of humans. The maintenance of masses of RPA bots is a typical problem that organizations with high numbers of RPA bots are addressing. Very often, organizations spend more time on maintenance than automating other processes (Kedziora and Penttinen, 2021; Průcha, 2021). Therefore, to facilitate the management of the RPA bot portfolio, it is advisable to have tools to detect problematic bots that need to be improved, repaired and/or redesigned.

In this research, we, thus, aim to validate assumption that measuring variability can provide valuable insights to management of RPA bots. We are searching for appropriate indicators that will objectively describe the behavior of RPA bots based on the variability of the processing time of a single case.

## 2 Related Work

Statistical process control (SPC) for detecting process performance and behavior is a relatively well-known activity. Many publications in operations research have already addressed this issue. SPC has crossed the boundary of the scientific environment extremely well and the same principles are also widely used in industry such as Six Sigma. The field of process mining focuses on performance, transparency, behavior and the root cause analysis of this behavior, variance analysis and other issues (van der Aalst, 2022). The evolution of the process mining field is progressing ensuring that process mining methods and research bring more value to organizations as they make decisions and take action based on these approaches (van der Aalst, 2016). Therefore, gaining an understanding of process behavior and acquiring insights to optimize automated processes emerge as an approach consistent with the current trend of process mining.

### 2.1 Variability and SPC in Process Mining

One of the goals of process mining is to convert data into valuable insights on which organizations can make decisions (van der Aalst, 2022). The quality and preparation of data for process mining is very time-consuming and, therefore, process mining should provide significant added value and companies should use the insights to take action (De Weerdt and Wynn, 2022; van der Aalst, 2016). Process mining in RPA processes is not very typical at the moment, however, it can be applied to automated processes for RPA (Egger et al., 2020; Průcha, 2021), where duration rates, bottlenecks and deviations can be identified. With advanced methods of process mining such as predictive process monitoring, enhancing the process data with data from other systems, root-cause analysis and online monitoring, it is possible to gain more valuable insights about the behavior of RPA processes including information why an RPA bot fails, whether the bot fails when executing a particular variant and whether RPA bots can keep up with the queue of cases (Burattin, 2022; Di Francescomarino and Ghidini, 2022; Fahland, 2022; Van Houdt and Martin, 2021).

The use of statistical methods in process mining represents the core of process mining. The most commonly used methods include root-cause detection, predictive process mining and explainable AI in process mining (Garcia et al., 2019; Jans et al., 2013; Tiwari et al., 2008; van der Aalst, 2016). Therefore, descriptive statistics and inferential statistics are very often used for process performance analysis to describe what is happening in the process effectively. Process variability is certainly one of the topics that resonates in process mining; for the query: "process variability in process mining", we obtained 2,250 results in WOS. In process mining, process variability can be separated into two basic divisions:

- whole process variability - how many variations exist
- the variability of a single variant or process - for example, how many runs of the process are consistent over time (Ayora et al., 2013).

In our research, we focus on the variability of one variant or process over time. Results regarding the whole process variability are not relevant to this study. The number of studies focusing on process variability in terms of duration is limited. There are 48 results in the WOS database for the query: variability in duration process mining and 129 results in the Scopus database for the query: variability in duration "process mining". Based on the relevance of the results, the query varied by quotation marks for each database. From the results obtained, as well as additional investigations, we could not find research that addresses process behavior based on duration, variability and/or robotic process automation.

A significant number of results were related to process variability in healthcare, where processes are typically variable by nature (Munoz-Gama et al., 2022). There is also a study that focuses specifically on process duration and uses process duration to investigate process behavior and process performance visualizing the results using a graph they have designed for this purpose called the performance spectrum (Denisov et al., 2018). Furthermore, there is research on clocking process time from a transistor-liquid crystal display to reduce variability and error behavior in manufacturing (Kang et al., 2016). These studies show that statistical methods can be used to analyze process behavior and reduce or detect problematic behavior.

## 2.2 Statistical Process Control

Statistical Process Control has been used primarily in manufacturing where the limits and requirements are rigidly defined (MacGregor and Kourti, 1995). In this environment, methods such as cumulative sum (CUMSUM), exponentially weighted mean (EWMA) and process capability work very well (Woodall, 2000). Despite the ingenuity and potential these techniques have, they always need initial values as the mean, i.e., a lower bound limit and an upper bound limit, to be used. In pharmaceuticals, chemistry, construction and engineering, where SPC has spread, very often there are limits and standards that are given (Stoumbos et al., 2000). In fields that do not have strict rules and standards, the use of SPC convex methods is more complicated.

Since there are fields where limits and norms are not strictly specified, such as most business processes, it is necessary to use other SPC methods such as variance statistics. There are approaches for using variance statistics for SPC such as comparing a multivariable process using a covariance matrix (Tang and Barnett, 1996a, 1996b). However, this approach reveals little about what is happening in the process and its behavior.

## 2.3 Measuring RPA performance

The most common key performance indicators (KPIs) for RPA bots have, from the start, been full time equivalent (FTE) hours saved and from these, the return on investment (ROI) of process automation has subsequently been calculated (Axmann et al., 2021; Wewerka and Reichert, 2020). In the early days of implementing RPA in organizations, this was also one of the main indicators for selecting processes for automation (Anagnoste, 2017; Willcocks et al., 2015b). With the advances in RPA, other benefits of robotic process automation have become apparent, such as mitigating process errors, providing 24/7 service, recording process steps for legislation, and increasing employee satisfaction with automation of routine activities (Aguirre and Rodriguez, 2017; Schuler and Gehring, 2018; Syed et al., 2020). With these advances, companies have begun to measure new KPIs such as run-time (velocity), cost per error, license utilization, exception rate, and average automation uptime (Aguirre and Rodriguez, 2017; Kokina and Blanchette, 2019; Blueprint, 2021; Syed et al., 2020; Teodorczuk, 2021; Wanner et al., 2019).

Views are also emerging that, in addition to the listed KPIs, firms should measure KPIs that are not common but provide additional value and decision support to organizations including break-fix cycle, break-fix person hours, break root causes, business value lost in downtime and consistency (Casey, 2019; Blueprint, 2021).

After a systematic review of the relevant literature, there arise two research questions which need to be answered:

RQ 1: Is it possible, based on the consistency/variability of the RPA bot, to determine the behavior and performance of the RPA bot?

RQ 2: Which indicator of variability is the most dependable for determining the performance of RPA bots?

## 3 Approach

In our research, we focused on answering the research questions mentioned above. For answering the research questions, we formulated two hypotheses.

For RQ1 we formulated the following hypotheses:

- Hypothesis 1: Variability is related to the performance of an RPA bot (success rate).
- Hypothesis 2: Outliers are related to the performance of an RPA bot (success rate).

For RQ2 we did not formulate any hypothesis, because we assumed in this research that the most dependable indicators are those with the strongest correlations.

For validation of the two hypotheses, we selected appropriate indicators of statistical dispersion (variability) which may faithfully describe the behavior of RPA bots based on the variability of the processing time of a single case.
We computed these values using real RPA processes where we had prepared processing times for each case processed by an RPA bot. The research process is displayed in Figure 1.

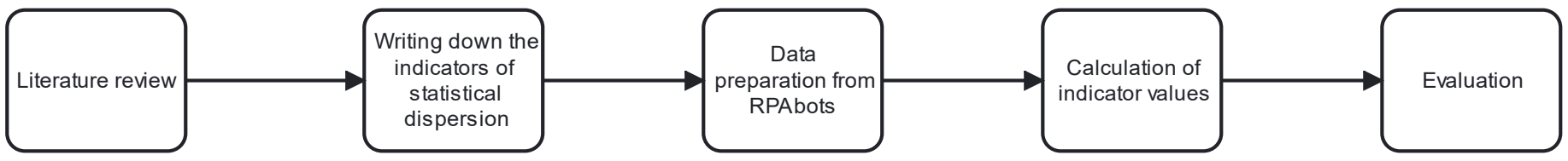

**Figure 1**: The research process

RPA processes were selected across various industries: 10 out of 12 processes are from the banking industry, where interval workflows are executed within the bank infrastructure. The rest of the RPA processes are customized automations of HR processes working with attendance systems. The selected processes are from two RPA platforms, namely Blue Prism and UiPath. Out of the 12 processes, we selected two benchmark processes to serve as examples for comparison. These are processes which work perfectly, while the other RPA processes are very problematic in that they exhibit extensive error behavior which requires repair. These two processes served as benchmarks in the comparison with the other processes. These two benchmark processes came from the same bank, and both used the same RPA platform version from the same distributor. The non-problematic processes have been in operation for more than 2 years and have undergone multiple optimizations. The problematic processes are after a roll out to production in the hypercare phase of an RPA. That is why is they are problematic; the production environment can sometimes cause processes to behave differently than during the development environment.
Basic information about the processes is presented in table 5. A more comprehensive description of the processes is prohibited by a non-disclosure agreement. The number of cases in table 5 is rounded to the nearest 10 to preserve anonymity.

**Table 1.** Basic data on benchmark processes

| Process | Success Rate (SR) | Outliers out of IQR (OoIQR) | Number of cases | Industry | Platform |
|---|---|---|---|---|---|
| EFP | 100 % | 0 % | 400 | Banking | Blue Prism |
| MP | 5 % | 0.09 % | 1000 | Banking | Blue Prism |
| P1 | 96 % | 8.18 % | 440 | Banking | Blue Prism |
| P2 | 93 % | 4.79 % | 150 | Banking | Blue Prism |
| P3 | 91 % | 27.7 % | 450 | Banking | Blue Prism |
| P4 | 19 % | 7.14 % | 90 | Banking | Blue Prism |
| P5 | 69 % | 25.19 % | 420 | HR process | Blue Prism |
| P6 | 82 % | 17.55 % | 380 | HR process | Blue Prism |
| P7 | 57 % | 9.52 % | 90 | Banking | UiPath |
| P8 | 94 % | 7.58 % | 1650 | Banking | UiPath |
| P9 | 99 % | 0.92 % | 310 | Banking | UiPath |
| P10 | 99 % | 17.99 % | 90 | Banking | UiPath |

### 3.1 Methodology for selecting indicators of statistical dispersion

There are surprisingly numerous methods for measuring variability and deviation from the norm. Probably the most common and widely used measures in statistics are: standard deviation, variance and coefficient of variation.

There is a significant difference between the results of these three coefficients. The standard deviation and variance always yield a result in absolute values, while the coefficient of variation yields a normalized result and can take on values from 0 to 1. Because RPA robots can take different amounts of time to process cases based on the number of steps in the process and the operations performed, the absolute values of duration change with it. It is not possible to compare two different processes based on standard deviation and variance. Hence, for our measurements, we will only select indicators that are normalized as a coefficient of variation. We show the difference between the results of absolute values and normalized values in Table 1.

**Table 2.** Difference between absolute value indicators and normalized indicators

| Indicator | Results for values (10, 20, 30) | Results for values (100, 200, 300) |
|---|---|---|
| Standard deviation | 8.16 | 81.65 |
| Variance | 66.67 | 6666.67 |
| Coefficient of variation | 0.41 | 0.41 |

### 3.2 Introduction of the indicators

To start with, Table 2 introduces concepts to understand the calculations of the selected indicators. The name is in the first column, the labeling in the second and the description is in the third column. Table 3 presents the selected indicators that have already been researched.

**Table 3.** List of values used for calculations

| Name | Labeling | Description |
|---|---|---|
| Standard Deviation | σ | Measure of the amount of variation of a set of values. (1) |
| Arithmetic Mean | μ | The central tendency of a set of numbers. (2) |
| Median | med | The middle value in a list ordered from smallest to largest. |
| Quantile – for a certain value | Q, $Q_{0.yz}$ | Characteristics of the data set and the quantile of a certain value, for example, Q0.95. |
| Value | x | A single value from the data set. |
| Number of values | n | The number of values in the data set. |
| Highest value | H | The highest value in the data set. |
| Lowest value | S | The smallest value in the data set. |
| Interquartile range | IQR | Describes the distribution of data between Q3 and Q1. (3) |
| Outlier | O | The data point that is significantly different from the others. |
| Outlier Count 1SD | Oc1SD | The number of outliers in the data set outside the distance of one standard deviation. (4) |
| Outlier Count IQR | OcIQR | The number of outliers outside Q1 - 1.5 * IQR and the number of outliers outside Q3 + 1.5 * IQR. (5) |
| Success Rate | SR | The percentage of successfully processed cases in the process. |

$$\sigma = \sqrt{\frac{\sum_{i=1}^{n} |x_i - \mu|^2}{n}} \quad (1)$$

$$\mu = \frac{1}{n} \sum_{i=1}^{n} x_i \quad (2)$$

$$IQR = Q_3 - Q_1 \quad (3)$$

$$Oc1SD = n < -SD \wedge n > SD \quad (4)$$

$$OcIQR = n < Q1 - 1.5 * IQR \wedge n > Q3 + 1.5 * IQR \quad (5)$$

**Table 4.** Selected indicators of statistical dispersion

| Name | Labeling | Description |
|---|---|---|
| Coefficient of variation | CV | CV is the ratio of the standard deviation and the arithmetic mean. (6) |
| Coefficient of range | CR | CR is the ratio of the difference between the highest and lowest values and the sum of the highest and lowest values. (7) |
| Coefficient of dispersion | CD | CD is the proportion of the deviation from the median and the median times 1/n. (8) |
| Coefficient of mean deviation | CMD | CMD is the ratio of the mean deviation and the arithmetic mean times 1/n. (9) |
| Coefficient between the quantile range | $CIQR_{90}$ | $CIQR_{90}$ is the proportion of the difference between Q0.90 and Q0.10 and the sum of Q¬¬0.90 and Q0.10. (10) |
| Gini Coefficient | GC | GC is defined as the mean of absolute differences between all pairs of individuals. To use the X formula, all values must be in ascending order. Calculation taken from (Kendall et al., 1994). (11) |
| Number of outliers out of one sigma | OoOS | OoOS is the ratio of the number of outliers outside a distance of one standard deviation to the total number of all cases. (12) |
| Number of outliers out of IQR | OoIQR | OoIQR is the proportion of OcIQR and the total number of all cases. (13) |

$$CV = \frac{\sigma}{\mu} \tag{6}$$

$$CR = \frac{H-S}{H+S} \tag{7}$$

$$CD = \frac{1}{n} * \frac{\sum_{i=1}^{n} |x_i - med|}{med} \tag{8}$$

$$CMD = \frac{1}{n} * \frac{\sum_{i=0}^{n} |x_i - \mu|}{\mu} \tag{9}$$

$$\mathrm{CIQR}_{90} = \frac{Q_{0.95} - Q_{0.05}}{Q_{0.95} + Q_{0.05}} \tag{10}$$

$$GC = \frac{\sum_{i=1}^{n} (2i-n-1)x_i}{n \sum_{i=1}^{n} x_i} \tag{11}$$

$$OoOS = \frac{\mathrm{Oc1SD}}{n} \tag{12}$$

$$OoIQR = \frac{\mathrm{OcIQR}}{n} \tag{13}$$

To find the number of outliers used in OoIQR and OcIQR, the calculation that is used is also applied to display the outliers in the boxplot. Together, Oc1SD and OcIQR are only the number of values that exceeded a defined threshold, so they are not individual values.

### 3.3 Methodology of calculation used

We have prepared the data for every single process, where we only worked with the times required to perform certain cases (time to complete one item in a process). Thus, we collected data for every process and all the cases performed by one RPA robot (process) and the time needed to accomplish one item was extracted for every single item. Each process has a different measured number of cases processed depending on the bot's workload. The case counts are always higher than 90 cases for a single RPA process, so the counts are sufficient for using descriptive statistical methods. A sample of the data can be seen in Table 4, where P1 and P2 are added outlier values. The data used are in seconds. The Python language and Jupyter notebooks were used for data analysis; also, a library[1] was programmed to calculate the above formulas. After calculating all the indicators, and in order to verify the accuracy of the prediction, we used the Pearson correlation coefficient to test the dependence between reality and the indicators.

**Table 5.** Sample data

| Process | Duration of one case in process (in seconds) |
|---|---|
| EFP | 213, 215, 210, 214, 211 |
| MP | 246, 238, 248, 235, 244 |
| P1 | 166, 173, 180, 182, 8 |
| P2 | 208, 199, 203, 496, 488 |

## 4 Results

In Table 6, there are values from indicators for every process we have data on with basic information such as the success rate (SR) and the percentage of outliers outside the IQR out of the total (OoIQR). The results in Table 6 are rounded to 2 decimal places except for the SR indicator, where the accuracy is only in whole numbers.

**Table 6.** Values of statistical dispersion indicators

| Process | CV | CR | CD | CMD | CIQR90 | GC | OoOS | SR | OoIQR |
|---|---|---|---|---|---|---|---|---|---|
| EFP | 0.29 | 0.81 | 0.25 | 0.24 | 0.35 | 0.16 | 0.33 | 100 % | 0 % |
| MP | 1.0 | 1.0 | 1.7 | 0.9 | 0.9 | 0.5 | 0.2 | 5 % | 0.09 % |
| P1 | 0.23 | 0.99 | 0.11 | 0.12 | 0.16 | 0.09 | 0.01 | 96 % | 8.18 % |
| P2 | 0.34 | 1.0 | 0.22 | 0.26 | 0.38 | 0.17 | 0.11 | 93 % | 4.79 % |
| P3 | 0.51 | 0.86 | 0.33 | 0.36 | 0.54 | 0.21 | 0.18 | 91 % | 27.7 % |
| P4 | 0.69 | 0.78 | 0.63 | 0.51 | 0.65 | 0.31 | 0.13 | 19 % | 7.14 % |
| P5 | 1.06 | 1.0 | 0.33 | 0.34 | 0.81 | 0.28 | 0.02 | 69 % | 25.19 % |
| P6 | 0.35 | 0.95 | 0.21 | 0.21 | 0.66 | 0.16 | 0.16 | 82 % | 17.55 % |
| P7 | 1.07 | 0.89 | 0.56 | 0.5 | 0.67 | 0.33 | 0.07 | 57 % | 9.52 % |

[1] The library can be found here: https://github.com/Scherifow/Dispersion-Statistic

| | | | | | | | | | |
|---|---|---|---|---|---|---|---|---|---|
| P8 | 0.2 | 0.8 | 0.2 | 0.2 | 0.3 | 0.1 | 0.2 | 94 % | 7.58 % |
| P9 | 0.11 | 0.92 | 0.06 | 0.06 | 0.07 | 0.04 | 0.02 | 99 % | 0.92 % |
| P10 | 0.24 | 0.98 | 0.06 | 0.11 | 0.94 | 0.06 | 0.06 | 99 % | 17.99 % |

We used a correlation matrix to test our assumptions about which indicator best predicts error behavior as well as the number of outliers. The correlation matrix can be found in Figure 2.

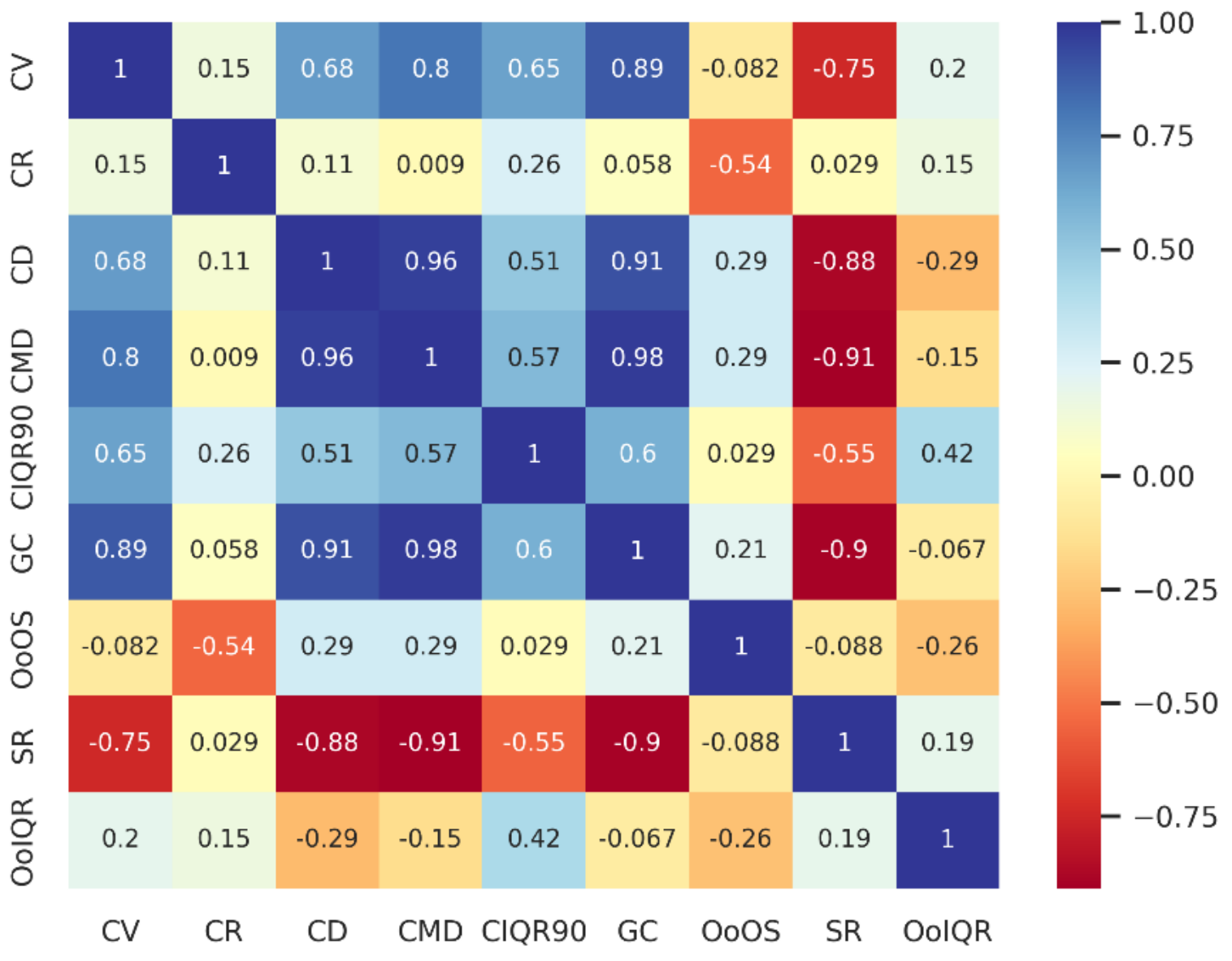

**Figure 2**: Correlation between reality and the indicator

The most interesting thing from the correlation matrix is that most of the variability indicators have a strong connection with the success rate; specifically, the Coefficient of variation, the coefficient of dispersion, the coefficient of mean deviation and the Gini coefficient all exhibited a direct correlation with the success rate. The correlation matrix clearly shows that the coefficient of mean deviation (CMD) has the strongest relationship with the success rate at -0.91. Coefficients between the quantile range of 90% only have a moderate dependency. The coefficient of range and the number of outliers out of one sigma show no dependency with the success rate. This could have been caused by the low sensibility of the indicators, and by the discovered finding related with Hypothesis two.

One interesting finding is related to Hypothesis two (H2). The success rate of an RPA process is not influenced by outliers, at least not in our data. So, if the indicators are influenced by outliers like CD and OoOS, the connection to the success rate is low.

At the beginning of this study, we proposed two research questions and hypotheses to answer the research questions:

RQ 1: Is it possible, based on the consistency/variability of an RPA bot, to determine the behavior and performance of an RPA bot?

- Hypothesis 1: Variability is to the performance of an RPA bot (success rate).
- Hypothesis 2: Outliers are related to the performance of an RPA bot (success rate).

RQ 2: Which indicator of variability is the most dependable for determining the performance of RPA bots?

Hypothesis 1: Based on our results, we can say that variability is related to the performance of an RPA.
Hypothesis 2: Based on our results, we were not able to prove that outliers are related to the performance of an RPA bot.

Thus, the answer for RQ1 is that we are able to determine the behavior of an RPA bot based on the variability of the process time.
The answer for RQ2 is that the coefficient of mean deviation (CMD) has the strongest relationship with the success rate. The correlation is very strong, so we can conclude that the CMD would be the most dependable indicator for determining the performance of RPA bots.

## 5 Discussion

In this research, we, thus, aimed to validate the assumption that measuring variability can provide valuable insights to management of RPA bots. We are searching for appropriate indicators that will faithfully describe the behavior of RPA bots based on the variability of the processing time of a single case.
Process variability is a common phenomenon and does not only occur in RPA processes, but is also common in industries where accuracy matters much more, such as, typically, chemistry, pharmacy and manufacturing (Munoz-Gama et al., 2022). There is also some variability in these industries, however, the variability is much smaller and this is due to the stringency and standards they are subject to (MacGregor and Kourti, 1995). High variability, many outliers, and a distribution of values that is flatter than the normal distribution most likely indicate some problem in any process regardless of the industry (Mapes et al., 2000; Munoz-Gama et al., 2022; van der Aalst, 2016). Over time, approaches to measure and remove this variability have been developed. Examples of such approaches include Statistical process control, process capability, six sigma methods, and some process mining methods.

Our research focused especially on variability development over time and more precisely, on the values of the duration of RPA bot case processing. Time variability is dealt with, for example, in manufacturing on production lines, where physical robots are used rather than software robots and it is necessary that all robots are synchronized to the production clock (Doyle Kent and Kopacek, 2021). Even though RPA bots are software and one would expect variability to be machine accurate, this is not the case. Since, quite often, additional variables enter into the processing that can prolong or complicate it. These complications can also occur in normal business processes. Typical examples might be bad input data, a bad form, or a bad format. Also, atypical variations of the process that are not as frequently dealt with include, for example, payment in a currency other than the default currency or processing data from within a foreign country or outside the country's economic space. Therefore, variability provides a good description of process behavior based on the input and the issues that affect it (Mapes et al., 2000; van der Aalst, 2016). Variability can also give a good indication of the maturity of a process and whether it is suitable for automation, thus, assisting in the selection of processes for automation. It can also serve in process selection screening for task-mining methods (Leno et al., 2021).

To refine the selection and description of process behavior, other statistical methods could be used including modified methods of statistical dispersion to increase sensitivity to certain behaviors. For example, the coefficient of variation is more sensitive to outliers than the coefficient of mean deviation. This leads to the fact that, even if there is a high success rate and there are no exceptions or errors in the process, the infrastructure where the robot is operating may be slow at certain times, resulting in outliers. CMD is not as sensitive to outliers and shows more of the main process flow, which can be an advantage if we are interested in that.

The results demonstrate that the outliers do not influence the success rate of RPA bots. However, if we are interested in what happens at the periphery and are concerned about the outlier values and not just the main flow, it might be advantageous to use other methods to detect outliers with statistical dispersion indicators more sensitive to outliers. Given the nature of RPA process behavior, it is likely that there will not be a high error rate in the data, i.e., neither a low success rate nor a high number of outliers, as confirmed by the correlation with the value of 0.19. The reason for this is that, if there is a high error rate, it is likely for the process to terminate simultaneously and, consequently, not many outliers will be generated. The method for identifying outliers used in the boxplot has some limitations. With a different outlier identification using, for example, a standard deviation within 2 standard deviations of the mean, where most values are found in a normal distribution, the result could be different.

Our research builds upon previous process behavior studies and presents ways to use statistical methods to detect process behavior. Indicating the possibilities for future research, it shows that it is possible to find other statistical methods to describe and detect process behavior. New ways to record process behavior and other characteristics such as performance will be beneficial for comparing processes and their performance (process benchmarking).

## 6 Conclusion

The results clearly show that, based on the indicators of statistical dispersion, the performance and behavior of RPA robots can be determined and, on this basis, the decision to perform an action can be made. This research gives RPA bot portfolio managers another tool to measure the performance of bots and identify bots for repair, which can lead to simplifying the maintenance of the entire RPA bot portfolio. Although the selected statistical dispersion indicators need not be limited to RPA processes, they can, nevertheless, be used in process-mining to detect and highlight malfunctioning processes in organizations. This can lead to a quicker analysis of problematic processes, or assist in the selection of an appropriate process for optimization. These insights can add further value to organizations and also extend the process mining toolkit with additional methods.

**Acknowledgments:** This research was made possible thanks to the Technical University of Liberec and the SGS grant number: SGS-2022-1004. This research was conducted with the support of Pointee.